\documentclass[10pt,twocolumn,letterpaper]{article}

\usepackage{iccv}
\usepackage{times}
\usepackage{epsfig}
\usepackage{graphicx}
\usepackage{amsmath}
\usepackage{amssymb}
\usepackage{booktabs}
\usepackage{array}

\usepackage[pagebackref,breaklinks,colorlinks]{hyperref}
\usepackage{xspace}
\usepackage{xcolor}
\usepackage{pifont}
\usepackage{authblk}
\newcommand{\xmark}{\ding{55}}

\renewcommand{\paragraph}[1]{\vspace{1.25mm}\noindent\textbf{#1}}

\def\ours{NTDM\xspace} 

\iccvfinalcopy 

\def\iccvPaperID{5281} 

\ificcvfinal\fi

\begin{document}

\title{Neural Textured Deformable Meshes for Robust Analysis-by-Synthesis}

\author[1]{Angtian Wang\textsuperscript{*}}
\author[1]{Wufei Ma\textsuperscript{*}}
\author[1]{Alan Yuille}
\author[2,3]{Adam Kortylewski}
\affil[1]{Johns Hopkins University}
\affil[2]{Max Planck Institute for Informatics}
\affil[3]{University of Freiburg}

\maketitle

\setcounter{footnote}{0}
\renewcommand{\thefootnote}{\fnsymbol{footnote}}
\footnotetext{* indicates equal contribution.}


\ificcvfinal\thispagestyle{empty}\fi

\begin{abstract}
Human vision demonstrates higher robustness than current AI algorithms under out-of-distribution scenarios.
It has been conjectured such robustness benefits from performing analysis-by-synthesis.
%
%
%
%
Our paper formulates triple vision tasks in a consistent manner using approximate analysis-by-synthesis by render-and-compare algorithms on neural features.
In this work, we introduce Neural Textured Deformable Meshes (\ours), which involve the object model with deformable geometry that allows optimization on both camera parameters and object geometries.
%
%
%
The deformable mesh is parameterized as a neural field, and covered by whole-surface neural texture maps, which are trained to have spatial discriminability.
During inference, we extract the feature map of the test image and subsequently optimize the 3D pose and shape parameters of our model using differentiable rendering to best reconstruct the target feature map.
%
%
We show that our analysis-by-synthesis is much more robust than conventional neural networks when evaluated on real-world images and even in challenging out-of-distribution scenarios, such as occlusion and domain shift.
Our algorithms are competitive with standard algorithms when tested on conventional performance measures. 
\end{abstract}

\section{Introduction}
Deep neural networks are typically designed to perform a single vision task and can achieve high performance on that task. 
However, humans are capable of performing multiple recognition tasks simultaneously and in a highly robust manner, \ie, generalizing under occlusion or environmental changes.
Cognitive studies suggest that the robustness of the human visual perception arises from the analysis-by-synthesis process \cite{neisser1967cognitive, yuille2006vision}. 
Current generative AI systems also employ the analysis-by-synthesis process by typically using a graphics pipeline, along with an explicit 3D representation of the object, such as a CAD model. These systems generate images of an object class and then search for the model parameters that best reconstruct a given test image.
Recent research has shown that performing render-and-compare algorithms on neural features can significantly improve the robustness of these systems under partial occlusion and domain shift \cite{wang2021nemo}.
%

%
\begin{figure}
    \centering
    \vspace{-2mm}
    \includegraphics[width=\linewidth]{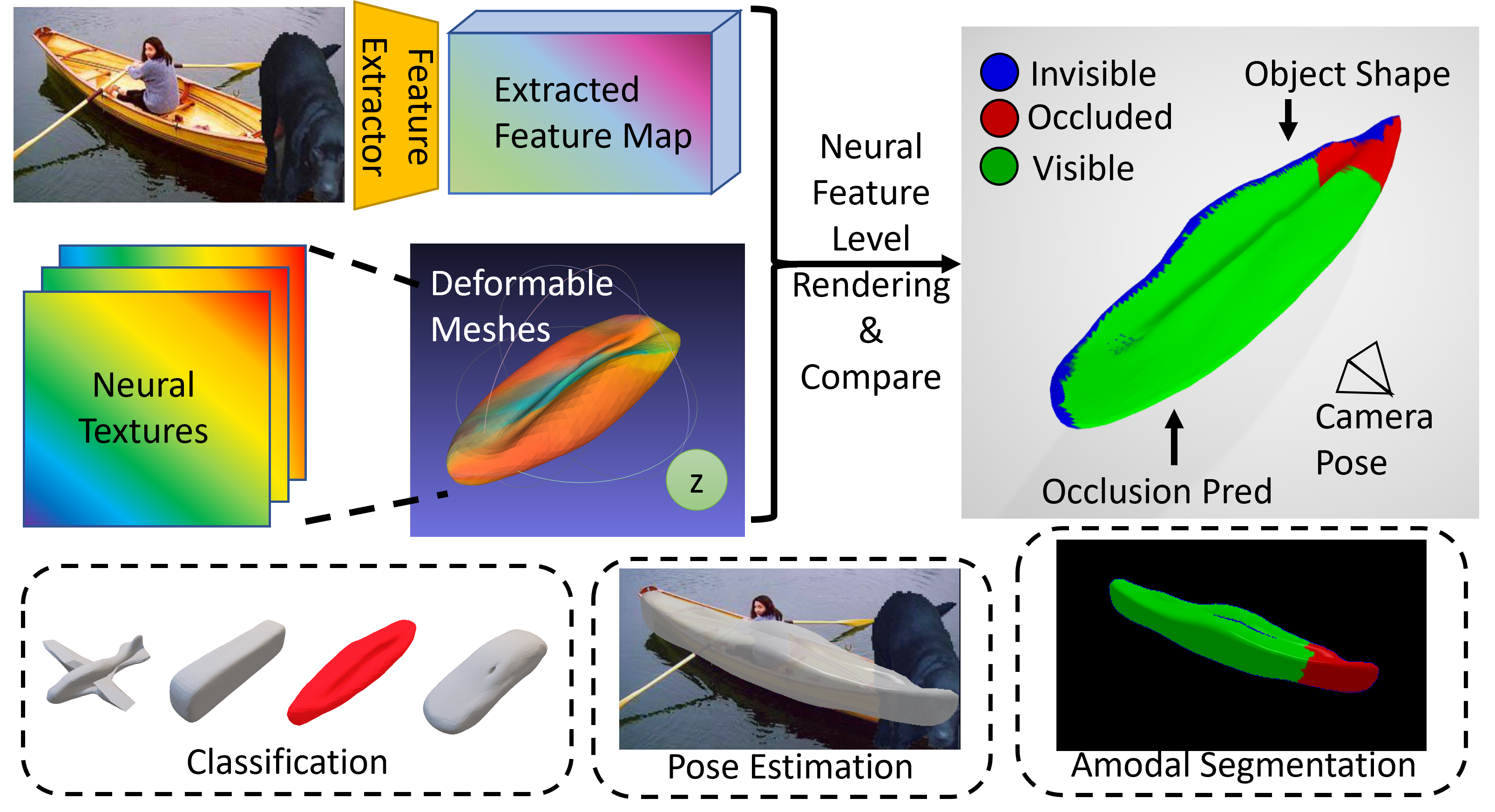}
    \vspace{-2mm}
    \caption{ \ours represents objects as category-level neural textured deformable meshes. For inference, we optimize camera pose, shape latent $z$, and object scale via gradient-based minimization of a reconstruction error between the extracted features and the rendered features. Using the optimized model parameters, \ours predicts object pose, shape, category, and occlusion jointly and consistently in a robust manner.}
    \label{fig:teaser}
    \vspace{-4mm}
\end{figure}

Another limitation of current computer vision algorithms is that they are unable to recognize objects comprehensively in the same way that humans can, without being constrained by a particular scope. 
For example, the algorithm's predictions are limited to specific tasks, unlike humans who can identify objects in a broader context.
While standard deep neural network approaches for multi-task learning typically involve adding multiple heads, where each head provides predictions for a specific task. However, this approach suffers from fundamental limitations, such as predictions for each task needing to be determined when designing network architecture, annotations for each task required for training, and lacking consistency in predictions from different branches \cite{zamir2020robust}.
On the other hand, analysis-by-synthesis approaches infer 3D scene parameters by reconstructing the input, which leads to a comprehensive recognition of the object 
with inherent consistency among vision tasks.

In this work, we introduce \ours, a 3D geometry-aware neural network architecture that implements an analysis-by-synthesis approach to computer vision, and hence is able to predict multiple visual recognition tasks in a unified manner, while also being exceptionally robust (Figure \ref{fig:teaser}).
%
Our model builds on and significantly extends recent work on generative models of neural network features. 
%
Specifically, we build on the concept of neural mesh models \cite{wang2021nemo,iwase2021repose} that represent objects as meshes and learn a generative model of the neural feature activations at each mesh vertex. 
These models solve vision tasks like pose estimation and 3D-aware image classification through a render-and-compare process.
The key advantage of performing render-and-compare on neural network features is that these can be trained to be invariant to instance-specific details, which makes the inference process efficient and robust.
%
The core limitation that prevents these methods from predicting other vision tasks, such as segmentation, is that they assume a fixed mesh geometry, which simplifies the learning and inference process but prevents them from estimating the object boundary accurately.

%
In order to perform analysis-by-synthesis for both object pose and geometry in an efficient manner, we introduce the concept of \textit{Deformable Meshes with Neural Textures} (\ours).
We present a framework for learning \ours and describe the inference process that enables the model to perform multi-task visual recognition in a unified manner.
At the core of our model is a deformable mesh geometry that is represented by a mesh template and a deformation field that is parameterized by a multi-layer perceptron (MLP) \cite{zhang2021ners} and trained from a few CAD models of the object class (typically 4-10 models).
Related works often represent deformable object geometries implicitly as a level-set in a volume of signed distances \cite{park2019deepsdf} or occupancies \cite{mescheder2019occupancy}. 
However, these representations are computationally expensive to render, while mesh representations in general can be rendered very efficiently and hence are preferable for render-and-compare approaches.
%
We model the appearance of the object as a neural texture map, which is trained in a discriminative manner to enhance the classification performance while also avoiding local optima in the reconstruction loss.
During inference, first a feature map is extracted using a CNN, and subsequently the 3D pose and deformation of the mesh is optimized via render-and-compare based on the reconstruction error between the rendered feature map and the target. 
After convergence, we perform image classification, pose estimation, and amodal segmentation using the optimized model parameters.

We evaluate \ours on the PASCAL3D+ \cite{pascal3dp}, the occluded-PASCAL3D+ \cite{wang2020robust} dataset, and the OOD-CV \cite{ood_cv} dataset, which was explicitly designed to evaluate out-of-distribution generalization in computer vision models.
%
Our experiments show that \ours performs competitively to all baselines while being highly robust in OOD scenarios.

In summary, our main contributions are:
\begin{itemize}
\item 
We introduce \ours, a neural network architecture that implements an analysis-by-synthesis approach and is hence able to perform multi-tasking robustly. 
\ours is composed of a deformable mesh geometry that is parameterized by a template mesh, a neural field of shape deformations, and a discriminatively trained neural texture. 
\item We demonstrate the versatility of our network architecture on a variety of datasets, where it performs competitively to single-task models while also being highly robust in out-of-distribution scenarios.
\end{itemize}
\section{Related Work}

\paragraph{Category-level pose estimation.} Category-level pose estimation estimates 3D orientations of objects in a certain category. A classical approach was to formulate pose estimation as a classification problem \cite{tulsiani2015viewpoints,mousavian20173d}. Subsequent works can be categorized into two groups, keypoint-based methods and render-and-compare methods. Keypoint-based methods \cite{pavlakos20176,zhou2018starmap} first detect semantic keypoints and then solve a Perspective-n-Point problem to find the optimal 3D pose. Render-and-compare methods \cite{wang2019normalized,chen2020category} predict the 3D pose by fitting a 3D rigid transformation to minimize a reconstruction loss. \cite{wang2021nemo,iwase2021repose} proposed feature-level render-and-compare that are invariant to intra-category nuisances and variations.

\paragraph{Amodal segmentation.} Amodal segmentation aims to predict the region of both visible and occluded parts of an object. Related works on amodal segmentation often adopt a fully-supervised approach, with training supervisions coming from human annotations \cite{follmann2019learning,qi2019amodal} or synthetic occlusions \cite{li2016amodal,ling2020variational,nguyen2021weakly}. Recent work \cite{sun2022amodal} introduces a Bayesian approach that is trained on non-occluded objects only and does not require any amodal supervision. Moreover, our model takes a 3D-aware approach for amodal segmentation such that our probabilistic model is built on top of deformable object meshes. As a result, our model does not require any amodal segmentation annotations but achieves more accurate boundaries compared to baseline models.

\paragraph{Multi-tasking.} Multi-task models are trained to solve multiple tasks simultaneously and are widely adopted in many areas \cite{collobert2008unified,deng2013new,ramsundar2015massively}. It has been found to generalize better by leveraging domain-specific information contained in training signals of related tasks \cite{caruana1997multitask} and be more parameter-efficient \cite{mahabadi2021parameter}. Previous works usually rely on a multi-head architecture \cite{He_2017_ICCV,li2020rtm3d} and despite models being supervised with auxiliary loss functions \cite{kendall2018multi,li2020rtm3d}, predictions from individual heads tend to be inconsistent, especially in out-of-distribution scenarios. Instead, we propose DMNT that substitutes multiple prediction heads with a 3D model of generative features and solves multiple tasks from a unified perspective.

\paragraph{Analysis-by-Synthesis.} Our work is built on feature-level render-and-compare \cite{wang2021nemo} which approximates the analysis-by-synthesis approach \cite{grenander1970unified,grenander1996elements} in computer vision. Analysis-by-synthesis approaches are found to enable efficient learning \cite{wang2021neural} and largely enhance robustness in out-of-distribution scenarios, such as partial occlusion \cite{kortylewski2021compositional,wang2021nemo,wang2020robust,ma2022robust} and out-of-distribution textures and shapes \cite{ood_cv}. Our DMNT extends the previous works with a deformable 3D representation of neural features that learns a more characteristic representation of the scene and allows the model to solve multiple tasks jointly.

\begin{figure*}
    \vspace{-3mm}
    \centering
    \includegraphics[width=0.95\linewidth]{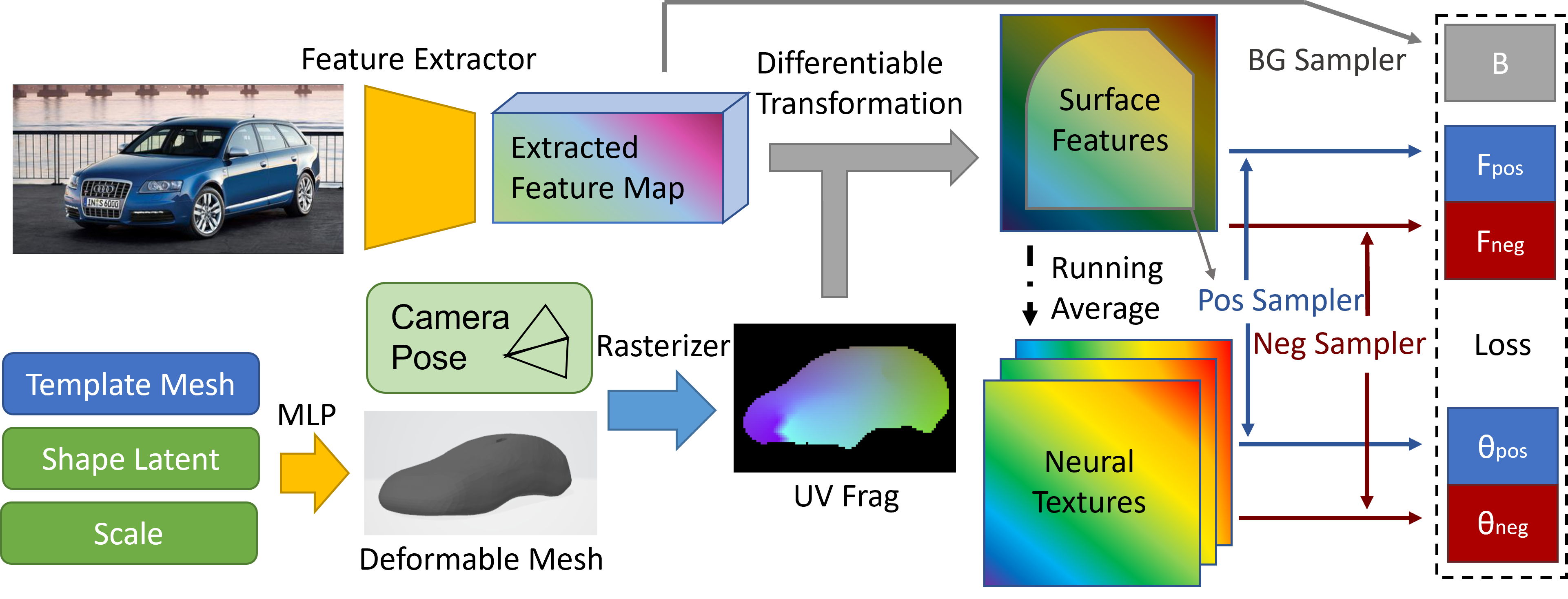}
    \vspace{-2mm}
    \caption{Training pipeline of \ours. We first extract features from the training image. Then we compute the deformable mesh via deforming a sphere template using an MLP with ground-turth latent $z$ inputs. A UV fragment is computed by rasterizing the deformable mesh under ground-truth camera. The surface features are computed via a differentiable transformation from image features given the UV fragments. Finally, we update the neural textures and compute the constrastive training loss via sampling positive and negative examples from image features for the surface features, neural textures, and background features.}
    \label{fig:training}
    \vspace{-1mm}
\end{figure*}

\section{Deformable Meshes with Neural Textures}
In the following, we first introduce the definition of our deformable mesh representation and the neural texture \ref{method:deformable}. Subsequently,  we define the probabilistic model of \ours \ref{method:model} and introduce the training \ref{method:train} and inference pipeline \ref{method:inference}.

\subsection{Deformable Meshes and Neural Textures}
\label{method:deformable}
We introduce \textbf{Deformable Meshes} $\Gamma$, which represent variable instances of an object category with a continuous deformation field on mesh vertices.
Specifically, given a sphere template mesh $\Upsilon$, the deformable mesh is defined as:  
\begin{equation}
\Gamma (z) = \{s \cdot (v + \Psi(v, z)) , v \in \Upsilon \}
\label{equ:dmesh}
\end{equation}
where $\Psi$ is an MLP that controls the mesh deformation \cite{zhang2021ners} via displacement of each vertex $v$, and $s=[s_h, s_w, s_d]$ is the average scale of the deformable mesh, which is optimized during the training. The latent variable $z$ controls the shape deformation of the mesh.


The mesh deformation network $\Psi$ is trained with a set of training meshes $\{ \Lambda_k \}$ , \ie, CAD models from the dataset. While the existing CAD models provide variable 3D geometries for each category, it is difficult to estimate the correspondence between them due to varying typologies and number of vertices. Thus, it is difficult to directly deform the CAD models via a common template. 
Instead, we propose to learn the correspondence among the provided meshes to build a deformable mesh representation. Specifically, we evaluate the distance between our deformable mesh with a specified latent $z_k$ and a target mesh $\Lambda_k$ via using the distance between vertex $v$ to the mesh faces $f$:
\begin{align}
    d(\Gamma (z_k), \Lambda_k) = \sum_{v \in \Gamma (z_k)} \min_{f \in \Lambda_k} d(v, f) + \sum_{v \in \Lambda_k} \min_{f \in \Gamma (z_k)} d(v, f)
\end{align}
where $z_k$ is a one-hot encoding vector of the index of the mesh $k$, \eg $z_1=[1, 0, ..., 0]$. 
To train the network $\Psi$ and $s$, we minimize $\sum _{k=1}^N d(\Gamma (z_k), \Lambda_k)$. We apply consistency constraints on the surface normals and a Laplacian smoothing loss \cite{nealen2006laplacian} on each $\Gamma (z_k)$ to regularize the shape (see supplementary for details). 


We define the \textbf{Neural Textures} $\Theta \in \mathbb{R}^{b \times q \times q \times d}$, where $q$ is the size of the feature map, $b$ is the number of viewing bins, on the surface of the mesh $\Gamma$, which are stored as square feature maps that contain feature vectors on each pixel $\theta_{b,u,v} \in \mathbb{R}^d$ for each viewing bin. We denote the mesh surface as $\mathcal{S}$.
The coordinate of $(u,v)$ is defined via the polar coordinates of locations on the sphere template mesh, such that for each point on the surface, we can easily compute its corresponding $(u, v)$ via Equation \ref{equ:dmesh}. 

To avoid local minima during optimization of the shape and pose, the Neural Textures are trained to learn 3D discriminative features, such that the distances in features space are correlated to the distance between their locations in the 3D space, \ie, features are similar to each other when near each other spatially, and vice versa. Additionally, the features are also learned to be spatially smooth via controlling the discriminability. The learned neural features provide a correct gradient on parameters when reconstructing the feature observations.

\begin{figure}[b]
    \centering
    \includegraphics[width=0.47\textwidth]{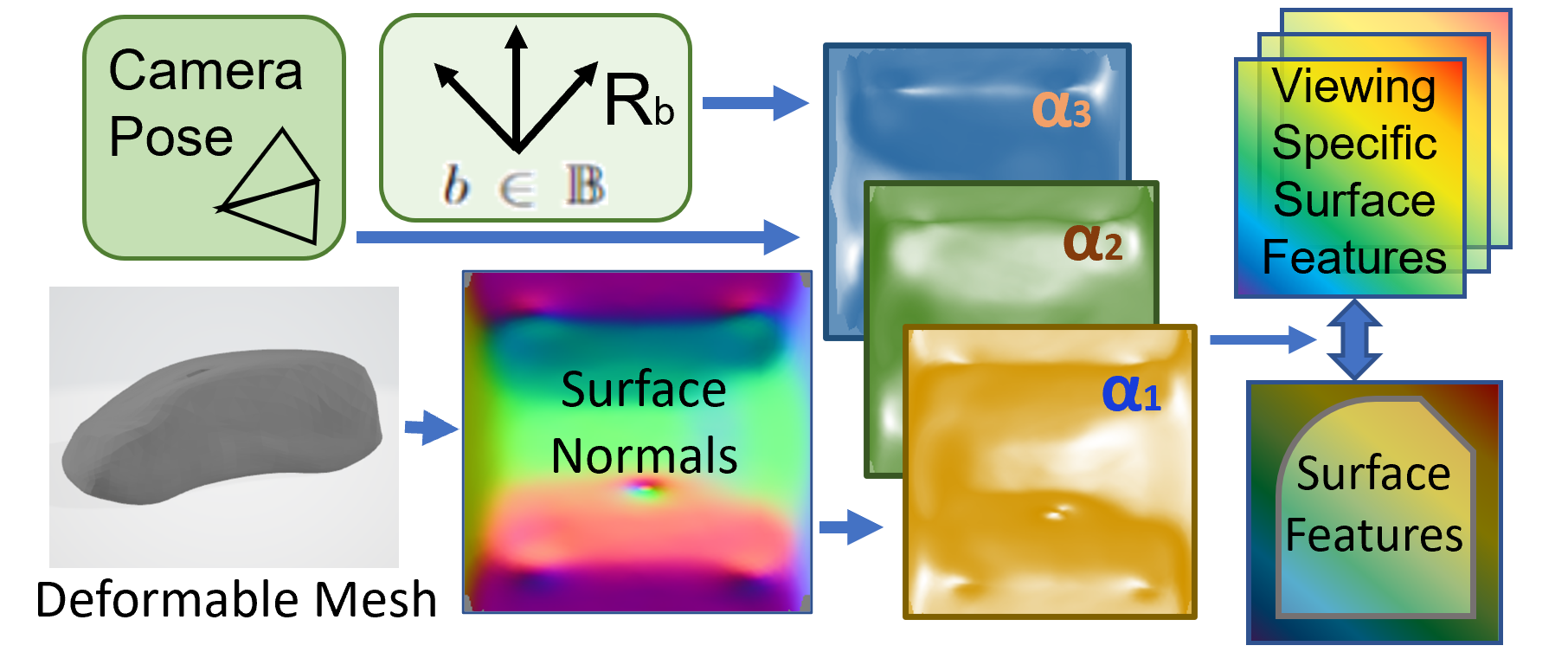}
    \caption{Conversion between viewing specific and non-specific surface features. Given a deformable mesh, we first compute the surface normals on mesh surface $\mathcal{S}$. Then we compute rotated surfaces by applying a set of rotations $R_b(\mathbf{n})$, and compute the viewing coefficients $\alpha_b$ with the dot product of $R_b(\mathbf{n})$ and viewing direction $d$. Viewing specific and non-specific surface features are converted via doting the coefficients $\alpha_b$.}
    \label{fig:vp}
\end{figure}

\subsection{Analysis-by-Synthesis via Rendering Neural Textures}
\label{method:model}
We formulate \ours as a probabilistic generative model of neural feature activations. 
Given an input image $I$, we extract the features via a convolutional neural networks
$\Phi(I) = F^l$.  
Then, we normalize the extracted feature activations $F=F^l / \Vert F^l \Vert$. We compute the object likelihood via:
\begin{equation}
    p(F|\Gamma, \Theta, c, m, B) = \prod_{i \in \mathcal{FG}} p(f_i|\Gamma, \Theta, c, m) \prod_{i \in \mathcal{BG}} p(f_i|B)
\end{equation}


where $\mathcal{FG}$ and $\mathcal{BG}$ denote the foreground and background, respectively, $i$ donates the pixel on the image plane, $m$ is the camera extrinsic parameters, $B$ is a set of feature vectors that model backgrounds.
The foreground feature likelihoods follow a Gaussian distribution:

\begin{align}
    p(f_i|\Gamma, \Theta, c, m) &= \sum_b \frac{\alpha_b}{\sigma\sqrt{2\pi}} \exp\left(-\frac{1}{2\sigma^2}\lVert f_i - \theta^{c}_{u, v,b}\rVert^2\right)
\end{align}

where $(u, v) \in \mathcal{S}$ is the corresponding location on the object surface that projects onto the pixel $i$ on the image plane. 
As Figure \ref{fig:vp} shows, $\alpha_b$ is the viewing coefficient:
\begin{align}
    \alpha_b=\frac{exp(T \cdot d \cdot  R_b(\mathbf{n}))}{\sum_b exp(T \cdot d \cdot  R_b(\mathbf{n}))}, {b} \in \mathbb{B}
    \label{equ:viewing}
\end{align}
where $T$ is a softmax temperature, $d$ is the normalized viewing direction from camera location to surface point, $\mathbb{B}$ is a set of rotations vectors $R_b$. Using each rotation vector, on each pixel $(u, v)$, we compute the direction with surface normals $d^{'}_{(u, v)} = R_b(\mathbf{n}_{(u, v)})$. In practice, $\mathbb{B}$ is a fixed set.

The background feature likelihoods are computed by:
\begin{align}\label{eq:back}    
    p(f_{i'}|B) &= \frac{1}{\sigma\sqrt{2\pi}} \exp\left(-\frac{1}{2\sigma^2}\lVert f_{i'} - \beta\rVert^2\right)
\end{align}
where $\beta \in \mathbb{R}^d$ is each features in $B$.


\subsection{Training \ours}
\label{method:train}
Figure \ref{fig:training} shows the training pipeline of \ours. In order to train the feature extractor and learn the neural texture jointly, we utilize the EM-type learning strategy introduced by CoKe\cite{bai2020coke}, which iteratively trains the feature extractor and updates the stored neural features.

We first obtain the normalized feature from the image $F=\Phi_W(I)$, where $W$ are the parameters in the feature extractor. 
Simultaneously, we compute the deformable mesh $\Gamma$ with a ground-truth $z$ and rasterize the mesh into the UV fragment $\mathrm{U} = \Re (m, \Gamma(z))$ under a ground-truth pose $m$. Then, we transform the features from the image plane into the surface features $f_{u_i, v_i} = f_i$ for all $i \in \mathrm{U}$. The transformation interpolates features in each pair of nearest pixels into quadrilateral regions onto the surface feature map, which also provides a mask of the visible region $\mathcal{V}$ of $\mathcal{S}$. 
Subsequently, we compute the view-specific features following Equation \ref{equ:viewing}: $f_{u_i, v_i, b} = \alpha_b \cdot f_{u_i, v_i}$. We update $\Theta$ with those features from visible regions $\mathcal{V}$ of the transformed feature map using the momentum update strategy \cite{bai2020coke}.


To learn the spatial discriminative features, we include a term that maximizes the log-likelihood of the generative model in the training loss:

\begin{align}
\begin{split}
    \mathcal{L}_{ML}(F,\Gamma, \Theta,m) 
    =-\ln p(F|\Gamma, \Theta, m) \\
        = \epsilon -\sum_{(u, v)\in \mathcal{S}} -\frac{1}{2\sigma^2}\lVert f^{c}_{u, v} - \theta^{b, c}_{u, v}\rVert^2 
\end{split}
\end{align}

where $\epsilon$ is a constant that $\epsilon=H \cdot W \cdot \left(\ln \frac{1}{\sigma_r\sqrt{2\pi}}\right)$, and $\theta^{c}_{u,v} = \sum_b \alpha_b \cdot \theta^{c}_{b,u,v}$.


We also compute a contrastive loss \cite{bai2020coke} to maximize the feature distance between features far from each other in the 3D space. 
To apply the loss, we first randomly sample a set of features from the visible part of the surface feature maps $f_{u,v}, (u,v) \in \mathcal{P} \subset \mathcal{V}$. Then, for each sampled $(u,v)$, we samples a set of points $(u^{'}, v^{'}) \in \mathcal{N} \subset \mathcal{S}$ that $\lVert (u^{'}, v^{'}) - (u, v) \rVert^2 > \tau$ as negative training examples, where $\tau$ is the threshold for controlling the spatial discriminability.
Then we compute a loss that maximizes the feature distance:
\begin{align*}
    \mathcal{L}_{Object}(F,\Theta_{\mathcal{S}}) &= - \sum_{(u, v) \in \mathcal{P}} \sum_{(u^{'}, v^{'}) \in \mathcal{N}} \lVert f^c_{u, v} - \theta^c_{u^{'}, v^{'}}\rVert^2
\end{align*}
To achieve the classification ability, for each image of $c \in \mathbf{C}$, we compute a loss to maximize the distance to a set features $\theta_{u', v'}, (u', v') \in \mathcal{M}$ from neural texture maps of other classes:
\begin{align*}
    \mathcal{L}_{Class}(F,\Theta^{\mathbf{C}}_{\mathcal{S}}) &= - \sum_{c \in \mathbf{C}, c' \neq c} \sum_{(u, v) \in \mathcal{P}} \sum_{(u^{'}, v^{'}) \in \mathcal{M}} \lVert f^c_{u, v} - \theta^{c'}_{u^{'}, v^{'}}\rVert^2
\end{align*}

Similarly, we retain a set of features $B = \{\beta_j\}$, which stores the negative examples from the background of images. This allows us to compute a loss that maximizes the objectness in contrast of the background:  
\begin{align}
    \mathcal{L}_{Back}(F, B) &= - \sum_{(u, v)\in \mathcal{P}}
    \sum_{j \in \mathcal{BG}} \lVert f^c_{u, v} - \beta^c_{j}\rVert^2 .
\end{align}

Then, the overall training loss is computed as:
\begin{align}
\begin{split}
    \mathcal{L}(F,\Gamma,\Theta,m,B) = \mathcal{L}_{ML}(F,\Gamma,\Theta,m) +  \mathcal{L}_{Object}(F,\Theta_{\mathcal{S}}) + \\ \mathcal{L}_{Class}(F,\Theta^{\mathbf{C}}_{\mathcal{S}}) + \mathcal{L}_{Back}(F, B)
    \label{equ:all_loss}
\end{split}
\end{align}
In terms of implementation, the overall loss is computed as a single cross entropy loss by concatenating all features of the negative examples together.

\begin{figure}
    \centering
    \vspace{-3mm}
    \includegraphics[width=\linewidth]{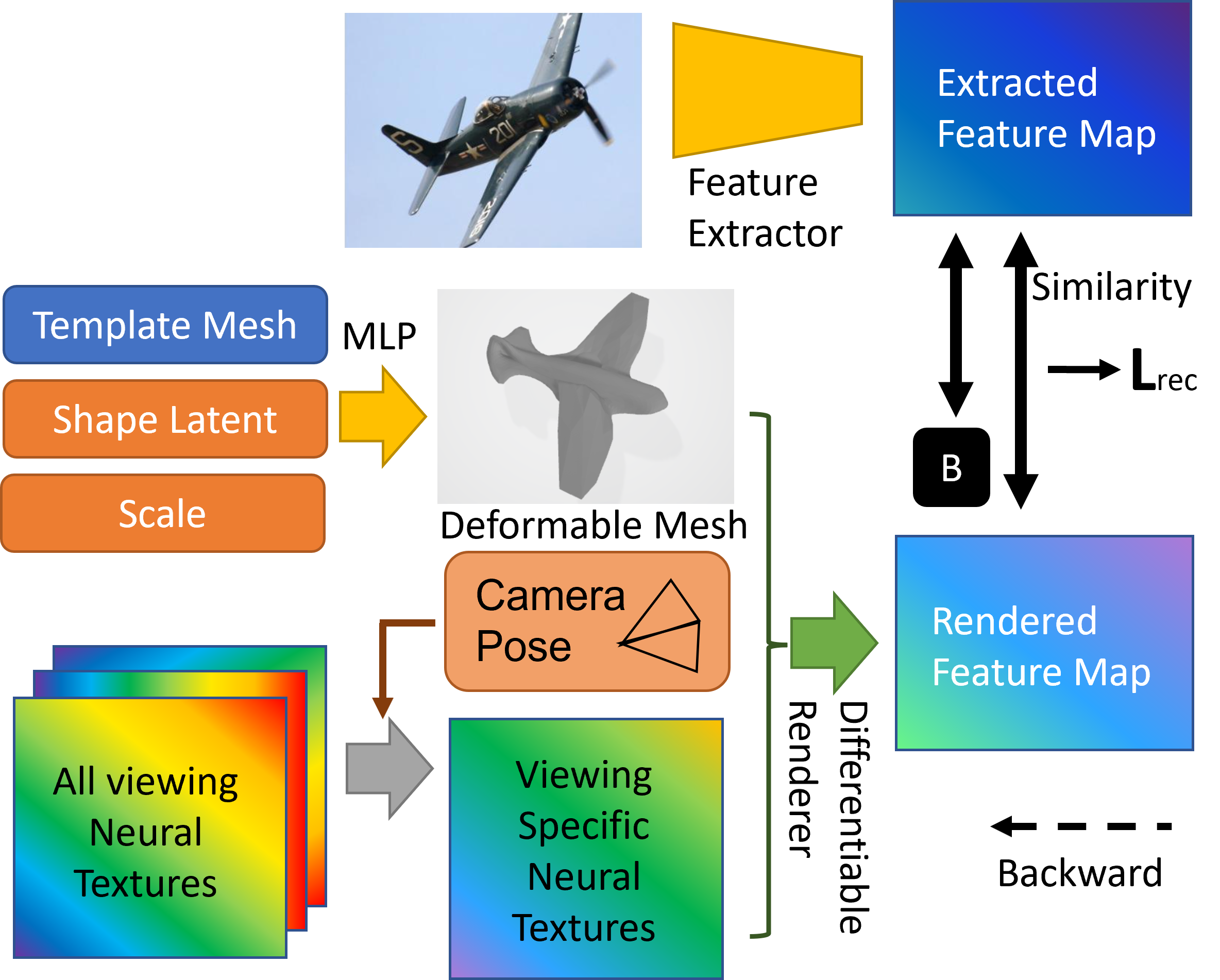}
    \caption{Inference pipeline of \ours. First, a feature map $F$ is extracted via the trained feature extractor. Next, the deformable mesh is initialized with a random latent and the average scale. Given a camera pose, we compute a viewing specific neural textures on mesh surface $\mathcal{S}$. Next, a differentiable renderer reconstructs the feature map $F^{'}$. Then \textcolor{orange}{Object scale}, \textcolor{orange}{latent} and \textcolor{orange}{camera pose} are jointly optimized via gradient descent that minimizes the reconstruction error between $F$ and $F^{'}$.}
    \label{fig:inference}
    \vspace{-4mm}
\end{figure}

\subsection{Multi-Tasking via Robust Optimization}
\label{method:inference}

Figure \ref{fig:inference} shows the inference pipeline of \ours. We first extract features via the trained feature extract from the image $F = \Phi(I)$. Then we conduct maximum likelihood estimation of $p(F| \Gamma(z), \Theta, m, B)$.
Specifically, given $k$ initialized latent $z_{init}$ (empirically we choose one-hot $z_{init}$ corresponding to each ground-truth subtype), and an object instance scale $s^{'}$ (initialized with $1$ in all direction), we compute the deformable mesh $\Gamma(z, s^{'})$. Using an initial camera pose $m=m_{init}$, we render the neural textured deformable mesh into a feature map $F'$. 
Then we conduct the foreground-background segmentation \cite{wang2021nemo} on all pixels covered by the projected object $\mathcal{O}$, which indicates if the pixel belongs to $\mathcal{FG}$ or $\mathcal{BG}$ by comparing the feature similarity $\sum_i f_i \cdot f{'}_i$ and $\sum_i f_i \cdot \beta$. Subsequently, we compute the feature reconstruction loss:

\begin{align}
\begin{split}
\mathcal{L}_{rec} &= 1 - \ln p(F| \Gamma(z, s'), \Theta, m, B) \\
&= 1 - (\sum_{i \in \mathcal{FG}} f_i * f{'}_i + \sum_{i \in \mathcal{BG}} f_i \cdot \beta)
\end{split}
\label{equ:rec}
\end{align}
We optimize camera pose $m$, shape latent $z$, and the object instance scale $s'$ via gradient to minimize the reconstruction loss. We use PyTorch3D \cite{ravi2020pytorch3d} to conduct the differentiable feature rendering and standard Adam optimizers \cite{kingma2014adam}. 


Once the optimization has converged, we obtain camera pose $m$ directly. The shape prediction is obtained by computing $\Gamma (z, s')$. The visible object segmentation is obtained from $\mathcal{FG}$, while the amodal segmentation is obtained via $\mathcal{O}$. For classification, we compute the reconstruction loss $\mathcal{L}_{rec}$ with neural textured meshes of all classes under the predicted parameters. We conduct classification  by finding the class with minimal reconstruction loss $\mathcal{L}_{rec}$.

 



\begin{figure*}
    \centering
    \vspace{-7mm}
    \includegraphics[width=\linewidth]{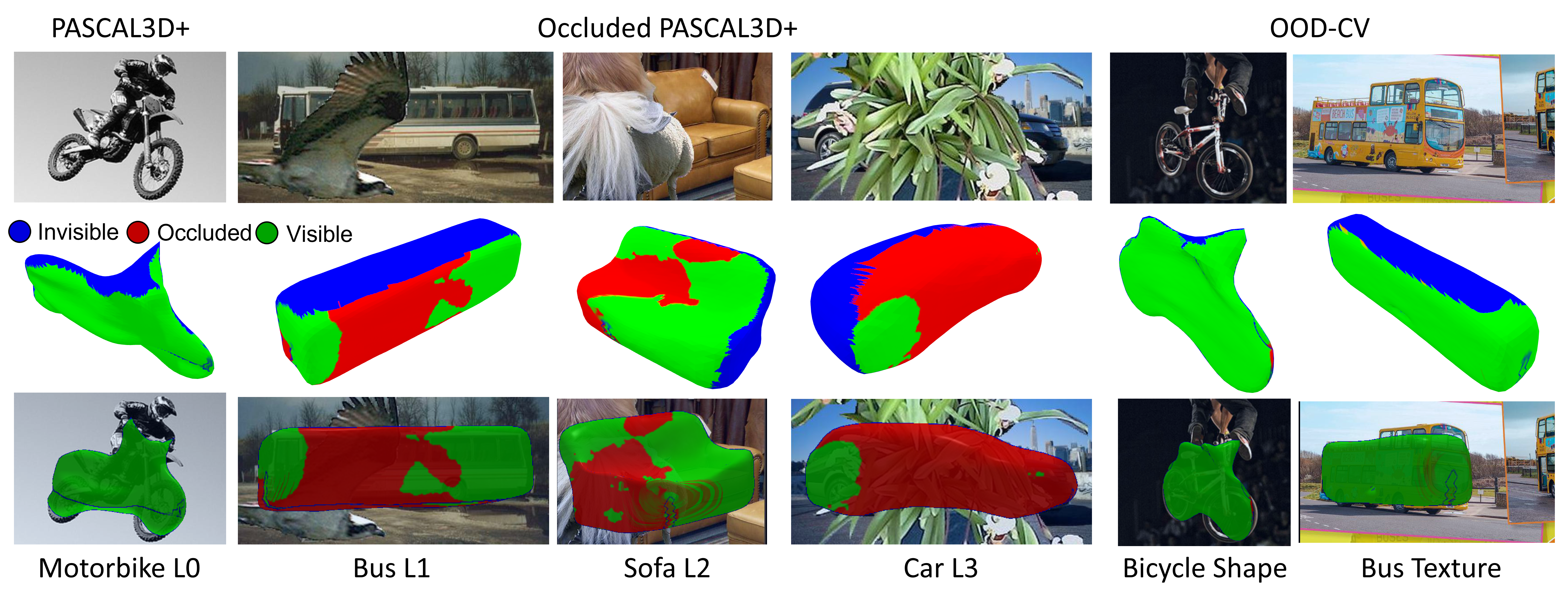}
    \vspace{-2mm}
    \caption{Visualizations for predictions of \ours on PASCAL3D+ (1st column), Occluded PASCAL3D+ (2nd-4th column) and OOD-CV (5th-6th column). The first row shows the input images. The second row visualizes the shape estimation of the deformable mesh, with colors indicates areas that are visible, invisible or occluded. For the third row, we render the predicted mesh under the predicted pose and superimposed it onto the input image. Note rendering process also gives the amodal segmentation predictions. }
    \label{fig:visualization}
    \vspace{-4mm}
\end{figure*}

\section{Experiments}

We evaluate the multi-tasking ability of \ours under I.I.D., which indicates the training and evaluation are under a same data distribution, and O.O.D., which evaluates on data out of the training distribution. We compare \ours with both task-specific approaches, \eg pose estimation (Res50 \cite{he2016deep}, StarMap \cite{zhou2018starmap}), robust pose estimation (NeMo \cite{wang2021nemo}), amodel segmentation (Bayesian \cite{kortylewski2021compositional}), and multi-task models (Multi-Task Mask R-CNN \cite{He_2017_ICCV}). Our results show that \ours achieves competitive performance on all tasks quantitatively, with a 3D interpretation of objects (figure \ref{fig:visualization}). 

\subsection{Experimental Setup}
We evaluate \ours and baselines on PASCAL3d+ \cite{pascal3dp}, Occluded PASCAL3d+ \cite{wang2020robust} and OOD-CV \cite{ood_cv} datasets. We follow the data pre-processing and training setup of NeMo \cite{wang2021nemo}.

\subsubsection{Datasets}

\paragraph{PASCAL3d+.} We evaluate 3D object pose estimation, amodal segmentation and classification of \ours and baselines methods on the PASCAL3D+ dataset \cite{pascal3dp}. The PASCAL3D+ dataset contains 11045 training images and 10812 validation images of 12 man-made object categories with class, segmentation and object pose annotations. PASCAL3D+ dataset also includes object CAD models. Note the CAD models are not accurately aligned to each object in image, but only give example of instances for each category (4-10 instance per category). We crop all training and evaluation images to center the object following NeMo.
 
\paragraph{Occluded PASCAL3d+.} The Occluded PASCAL3d+ \cite{wang2020robust} dataset is an extension of the PASCAL3D+ with man-made occlusion, which is created by superimposing occluders collected from the MS-COCO dataset onto objects in PASCAL3D+. In our experiment, we evaluate on three occlusion levels with increasing occlusion noted as L1 to L3.
 
\paragraph{OOD-CV.} The OOD-CV dataset \cite{ood_cv} is a benchmark introduced to evaluate model robustness in out-of-distribution scenarios. It includes O.O.D. examples of 10 categories that cover unseen variations of nuisances including pose, shape, texture, context, and weather.


\begin{table*}[t]
    \centering
    \resizebox{\textwidth}{!}{%
    \begin{tabular}{l|cccc|cccc|ccc|cccc}
        \toprule
        & \multicolumn{4}{c|}{Pose Estimation ACC${}_\frac{\pi}{6} \uparrow$} & \multicolumn{4}{c|}{Pose Estimation ACC${}_\frac{\pi}{18} \uparrow$} & \multicolumn{3}{c|}{Amodal Segm IoU $\uparrow$} & \multicolumn{4}{c}{Classification ACC (\%) $\uparrow$} \\ 
        Occlusion Level & L0 & L1 & L2 & L3 & L0 & L1 & L2 & L3 & L1 & L2 & L3 & L0 & L1 & L2 & L3 \\
        \midrule
        Res50-Pose & 88.1 & 70.4 & 52.8 & 37.8  & 44.6 & 25.3 & 14.5 & 6.7 & \xmark & \xmark & \xmark & \xmark & \xmark & \xmark & \xmark \\
        StarMap \cite{zhou2018starmap} & 89.4 & 71.1 & 47.2 & 22.9 & 59.5 & 34.4 & 13.9 & 3.7 & \xmark & \xmark & \xmark & \xmark & \xmark & \xmark & \xmark \\
        NeMo \cite{wang2021nemo} & 86.1 & 76.0 & 63.9 & 46.8 & 61.0 & 46.3 & 32.0 & 17.1 & \xmark & \xmark & \xmark & \xmark & \xmark & \xmark & \xmark \\
        Bayesian \cite{sun2022amodal} & \xmark & \xmark & \xmark & \xmark & \xmark & \xmark & \xmark & \xmark & 59.4 & 55.4 & 47.6 & \xmark & \xmark & \xmark & \xmark \\
        Res50-Class & \xmark & \xmark & \xmark & \xmark & \xmark & \xmark & \xmark & \xmark & \xmark & \xmark & \xmark & 98.7 & 91.6 & 74.1 & 42.6 \\
        MT Mask R-CNN & 85.0 & 66.6 & 56.2 & 47.3 & 46.1 & 29.9 & 15.6 & 5.9 & 66.6 & 56.2 & 47.3 & 95.7 & 78.7 & 52.8 & 25.6 \\
        DMNT & 86.4 & 74.8 & 61.0  & 40.2  & 61.3 & 44.8  & 30.1  & 13.7  & 67.9 & 63.5 & 57.5 & 94.1 & 85.0 & 67.8 & 43.2 \\
        \bottomrule
    \end{tabular}}
    \vspace{0.2mm}
    \caption{Comparison of multi-task performance on PASCAL3D+ dataset \cite{pascal3dp} (L0) and Occluded PASCAL3D+ dataset \cite{wang2020robust} (L1-L3).}
    \label{tab:exp-multi}
    \vspace{-2mm}
\end{table*}

\subsubsection{Implementation Details}

\ours uses the PyTorch3d \cite{ravi2020pytorch3d} rasterizer to infer the UV fragments which indicate the 3D correspondence from the image coordinates to the object surface. We implement the feature transformation using CUDA and develop a PyTorch API as a differentiable function, which computes the gradient not only toward the features but also to the UV fragments. Note this function could also be used for differentiable texture extraction beyond our current work. For implementation details, please refer to the supplementary. For both training and inference, we use the Perspective camera with a fixed focal length. The template mesh is a geodesic sphere with 2562 vertices.

\paragraph{Training.} We train \ours on the PASCAL3D+ dataset for 800 epochs with Adam Optimizer and an exponential learning rate starting from $10^{-4}$. \ours use the same feature extractor as NeMo, which is a ResNet50 with two additional upsampling layers. The Neural Textures contain Feature Maps of 7 viewing bins, which are computed by rotation vectors with 60-degree angular distance from each other. Each feature map has a resolution of $256 \times 256$ with $128$ channels. We update the Neural Textures Maps with a $0.9$ momentum. During training, we use a positive sampler with 1000 selections and a negative sampler with 2000 selections on each training example. Note we don't include any data augmentation in the training process.

\paragraph{Inference.}
We use the differentiable rasterizer to infer the UV fragment and use the grid sample function to convert Neural Textures into feature maps since the image features have a lower resolution compared to neural textures. Following NeMo \cite{wang2021nemo}, to speed up the inference process, we initialize the optimization process with 144 different camera poses (12 azimuths, 4 elevations, 3 in-plane rotations) and latent. We compute the reconstruction loss for each initialized combination and pick the one with the minimum loss as the starting point of optimization. Then we update all parameters with an Adam optimizer for 300 epochs with $lr=0.05$. The average inference time for pose and shape joint estimation on each image takes 12s on a single GPU. 

\subsubsection{Baselines}

\ours learns a neural representation of 3D deformable meshes and can simultaneously predict object classes, 3D poses, and object boundaries. In experiments, we compare our model with baselines from individual tasks, as well as a multi-task extension of a multi-head deep neural network.

\paragraph{3D pose estimation.} We compare \ours with StarMap \cite{zhou2018starmap}, NeMo \cite{wang2021nemo}, as well as standard deep neural network classifiers, ResNet-50 \cite{he2016deep}, that formulate pose estimation as a classification problem. StarMap is a keypoint-based approach, and NeMo learns contrastive features for render-and-compare. We follow the implementations in \cite{zhou2018starmap,wang2021nemo} to train the ResNet-50 pose estimation model.

\paragraph{Amodal segmentation.} We compare our model with Bayesian-Amodal \cite{sun2022amodal}, which extends deep neural networks with a Bayesian generative model of neural features. 
We use their official implementations to train on all categories in PASCAL3D+ dataset and evaluate on Occluded PASCAL3D+ dataset.

\paragraph{Classification.} We also train a standard ResNet-50 as  classification baseline using the PyTorch official version \cite{NEURIPS2019_9015}.

\paragraph{Multi-task deep neural network.} To compare our model with traditional multi-head network architectures, we extend a Mask R-CNN model \cite{He_2017_ICCV} with a pose estimation head that formulates object pose estimation as a classification problem. The model is end-to-end trained with ground-truth annotations including object classes, 3D poses, and object masks produced by known 3D meshes. \textcolor{black}{For more implementation details refer to the supplementary materials.}


\subsection{Multi-tasking in I.I.D. scenarios} \label{sec:exp-iid}

We are going through all tasks one by one in the following, but note that in contrast to most of our baselines, our model does all tasks jointly.

\paragraph{Pose Estimation.} 
The 3D object pose is defined via three rotation parameters (azimuth, elevation, in-plane rotation) of the viewing camera. Following previous works \cite{zhou2018starmap,wang2021nemo}, we evaluate the error between the predicted rotation matrix and the ground truth rotation matrix: $\Delta\left(R_{pred}, R_{gt}\right)=\frac{\left\|\log m\left(R_{pred}^{T} R_{gt}\right)\right\|_{F}}{\sqrt{2}}$. We report the accuracy of the pose estimation under given thresholds, $\frac{\pi}{6}$ and $\frac{\pi}{18}$.

\paragraph{Amodal Segmentation.}
Amodal segmentation predicts the region of both the visible and occluded parts of an object. Following previous works \cite{qi2019amodal,sun2022amodal}, we evaluate the average IoU between the predicted segmentation masks and the groundtruth segmentation masks.

\paragraph{Image Classification.} %
We evaluate the classification ability of both \ours and baselines. We report the top-1 accuracy between ground-truth class labels and predictions.

\paragraph{Results.} Table \ref{tab:exp-multi} show the multi-task performance for both \ours and baseline approaches. For the I.I.D setup, \ours achieve comparative performance compared to the single task approaches and significantly better pose estimation ability compared to multi-tasking Mask R-CNN. \ours achieves the \textbf{highest Pose accuracy under $\frac{\pi}{18}$}, which may benefit from the accuracy geometry compare to NeMo that uses cuboids as 3D geometry representation. Figure \ref{fig:visualization} shows both qualitative results of the pose estimation and segmentation, along with a 3D interpretation of the object produced by our model and visualized as a colored mesh. To obtain this colored mesh, we first conduct the $\mathcal{FG}$ and $\mathcal{BG}$ segmentation on the original image. We also compute the deformable mesh with the predicted latent $z$ and render it under the predicted camera pose. Using our introduced transformation function, we transform the occlusion segmentation onto the surface of the mesh, and fill the uncovered areas as invisible. Finally, we convert the segmentation into an RGB map and save the colored deformable mesh as a standard textured mesh file (.obj). This visualization demonstrates that \ours can produce a \textbf{comprehensive 3D understanding} of the object in a human interpretable way, which makes it feasible for more downstream tasks with \ours's pipeline.

\subsection{Robustness in O.O.D. scenarios}

\begin{table}[t]
    \small
    \centering
    \resizebox{\columnwidth}{!}{%
    \begin{tabular}{l|ccccc|c}
        \toprule
        Task \& Metric& \multicolumn{6}{c}{Pose Estimation ACC${}_\frac{\pi}{6} \uparrow$} \\
        Nuisance & shape & pose & texture & context & weather & mean \\
        \midrule
        Res50-Pose & 50.5 & 34.5 & 61.6 & 57.8 & 60.0 & 51.8 \\
        NeMo \cite{wang2021nemo} & 49.6 & 35.5 & 57.5 & 50.3 & 52.3 & 48.1 \\
        MT Mask R-CNN & 40.3 & 18.6 & 53.3 & 43.6 & 47.7 & 39.4 \\
        DMNT & 51.5 & 38.0 & 56.8 & 52.4 & 54.5 & 50.0 \\
        \midrule \midrule
        Task \& Metric& \multicolumn{6}{c}{Pose Estimation ACC${}_\frac{\pi}{18} \uparrow$} \\
        Nuisance & shape & pose & texture & context & weather & mean \\
        \midrule
        Res50-Pose & 15.7 & 12.6 & 22.3 & 15.5 & 23.4 & 18.1 \\
        NeMo \cite{wang2021nemo} & 19.3 & 7.1 & 33.6 & 21.5 & 30.3 & 21.7 \\
        MT Mask R-CNN & 15.6 & 1.6 & 24.3 & 13.8 & 22.9 & 15.3 \\
        DMNT & 20.7 & 12.6 & 32.6 & 16.6 & 33.5 & 23.6 \\
        \midrule \midrule
        Task \& Metric & \multicolumn{6}{c}{Amodal Segmentation IoU $\uparrow$} \\
        Nuisance & shape & pose & texture & context & weather & mean \\
        \midrule
        Res50-Pose & \xmark & \xmark & \xmark & \xmark & \xmark & \xmark \\
        NeMo \cite{wang2021nemo} & \xmark & \xmark & \xmark & \xmark & \xmark & \xmark \\
        MT Mask R-CNN & 46.6 & 44.5 & 51.3 & 44.7 & 46.0 & 46.3 \\
        DMNT & 48.0 & 48.9 & 54.7 & 49.4 & 53.8 & 51.0 \\
        \bottomrule
    \end{tabular}}
    \vspace{0.2mm}
    \caption{Comparison of multi-task performance -- pose estimation and amodal segmentation on OOD-CV dataset \cite{ood_cv}.}
    \vspace{-1mm}
    \label{tab:exp-ood}
\end{table}

We adopt the same evaluation protocol in Section \ref{sec:exp-iid} and evaluate multi-task performance on out-of-distribution scenarios in Occluded PASCAL3D+ and OOD-CV dataset.

\paragraph{Results under occlusion.} Table \ref{tab:exp-multi} shows the results of DMNT and baseline models on pose estimation, amodal segmentation, and image classification under occlusion. As we can see, \ours achieves comparable or better performance compared to the state-of-the-art task specific models, while \textbf{significant outperforms multi-task Mask R-CNN}. Regarding pose estimation, DMNT outperforms regression-based baselines and StarMap, and achieves comparable performance with NeMo under occlusion. Moreover, with a deformable mesh of the object on top of the feature backbone, DMNT solves object boundaries from a holistic perspective and outperforms all amodal segmentation baselines by a wide margin. \ours also perform object classification in a robust manner under partial occlusion.

\paragraph{Results under domain shifts.} We evaluate the pose estimation and amodal segmentation performance on OOD-CV and investigate the robustness under domain shifts -- shape, pose, texture, context, and weather. From Table \ref{tab:exp-ood}, we can see \ours achieves comparable pose estimation performance under $\pi/6$ accuracy and \textbf{outperforms all state-of-the-art models} when evaluating with a finer $\pi/18$ accuracy. Regarding amodal segmentation, \ours also outperform multi-task Mask R-CNN. We suggest that the deformable mesh and the spatial discriminative features learned by DMNT can adapt well to domain shifts, \eg shape and pose, and potentially useful in downstream tasks that requires highly robustness.

\begin{table}[t]
    \small
    \centering
    \begin{tabular}{l|cccc|c}
        \toprule
        Setup & Pose $\frac{\pi}{6}$ & Pose $\frac{\pi}{18}$ & Seg IoU \\
        \midrule\midrule
        full \ours & 91.4 & 60.5 & 80.7 \\
        w/o $\mathcal{BG}$ & 89.5 & 58.2 & 80.9 \\
        single $\alpha_b$ & 90.0 & 58.3 & 80.1 \\
        \midrule
        fix shape & 88.8 & 56.5 & 75.3 \\
        \midrule
        random & 90.1 & 59.4 & 80.4 \\
        average shape $1/n$ & 90.2 & 59.7 & 81.2 \\
        \bottomrule
    \end{tabular}
    \vspace{0.4mm}
    \caption{Ablation study on three object categories (sofa, bus, motorbike) from PASCAL3D+ dataset. We evaluate pose estimation accuracy and amodal segmentation IoU.}
    \label{tab:ablation}
    \vspace{-2mm}
\end{table}

\subsection{Ablation Study}
As Table \ref{tab:ablation} shows, we evaluate the contribution of each proposed component. Specifically, we evaluate the model on three categories (sofa, bus, motorbike) from the PASCAL3D+ dataset. The \textit{w/o $\mathcal{BG}$} setup indicates we only consider pixels in $\mathcal{FG}$ in reconstruction loss (equation \ref{equ:rec}). In the \textit{single $\alpha_b$} setup, we use only one viewing bin which keeps the $d_b$ along the surface normal direction. In this case, there is only a single neural texture map for each deformable mesh. For \textit{fix shape}, we use the average shape of deformable mesh to conduct inference, \ie, fixing latent as the average latent. \textit{random} and \textit{average} show the result using different initialization of object shapes latent during inference, \ie \textit{random} initialized with a random vector, \textit{average} initialized with the average object shape. The experiment demonstrates that all introduced components contribute to the final performance.

\section{Conclusion}
In this work, we propose \ours, which conducts \textbf{multiple vision task simultaneously in a consistent and robust manner}.
The core idea of \ours is the neural textured deformable meshes, which conducts gradient-based optimization of the shape and scale of the object, as well as the camera parameters simultaneously via neural feature level analysis-by-synthesis.
We introduce the learning pipeline for \ours that learn deformable meshes, neural textures, and feature extractor together so that components can cooperate with each other to enhance the performance. 
Experiments demonstrate that \ours produces a competitive performance compared to the task-specific approaches, and \textbf{extraordinary robustness} under occlusion and O.O.D. scenarios. Besides, we show that the predictions (shape, pose, occlusion) of \ours can be visualized as colored meshes, which makes the decision process of \ours interpretable and understandable.
Due to time and space limitations, we are not able to explore \ours in more downstream tasks. However, benefits from the generalization ability of the deformable 3D representation, \ours can be easily extended to more vision tasks, \eg part segmentation.



{\small
\bibliographystyle{ieee_fullname}
\bibliography{egbib}
}

\newpage
\appendix

\section{Implementation Details}

\subsection{Differentiable Transformation Function}

The differentiable transformation function takes inputs of an UV fragment $\mathcal{U}$ (which is obtained via interpolate the UV values of vertices using the barycentric weight from rasterization) and a feature $F$ on the image coordinate, and output the surface features $\hat{F}$ with a visibility mask $\mathcal{V}$. The transformation compute gradient to both $\mathcal{U}$ and $F$.

\begin{figure}[t]
\centering
\includegraphics[width=0.46\textwidth]{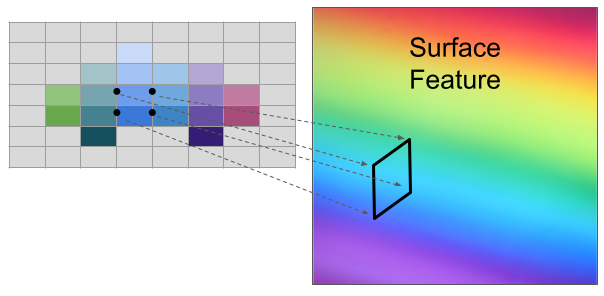}
\caption{Each four adjacent pixel in the UV fragment is interpolated as a quadrilateral with barycentric coordinates.}
\label{fig:feature_trans}
\end{figure}

\textbf{Forward. } As Figure \ref{fig:feature_trans} shows, for each four adjacent pixels $\mathcal{P} = (p_1, p_2, p_3, p_4)$ pairs on the $\mathcal{P} \in \mathcal{U}$, we first check if all of them are on the object. For each on-object pixel pair, we find the corresponded quadrilateral on the surface $\mathcal{S}$. Then we compute the barycentric coordinates inside the quadrilateral, which gives four weights $w_1, w_2, w_3, w_4$ on each surface pixel. Then we weighted sum the four feature vector on $p_1, p_2, p_3, p_4$ to compute the final value on the output surface feature:
\begin{align}
    F_{(u, v)} = w_1 \cdot F_{p_1} + w_2 \cdot F_{p_2} + w_3 \cdot F_{p_3} + w_4 \cdot F_{p_4}
\end{align}
In our implementation, for those cases that a pixel on surface is covered by multiple quadrilateral, we take a average of the value via store the total weight per surface pixels $\hat{w}_{(u, v)} = \sum_{ \mathcal{P} \in \mathcal{U} }\sum_k^4 w_k$. Then the visibility mask is computed as the area that $\hat{w}_{(u, v)} > 0$.
Also, in order to get rid of the quadrilateral cross the left to right or top to bottom boundary, we skip the quadrilateral larger than a threshold (set to be 0.2 of the surface size). 

\textbf{Backward. } In the backward process, assume the final loss is computed as $L$ and the up stream loss to the layer is $\frac{\partial L}{\partial F_{(u, v)}}$ on each surface pixel. The overall loss to the input feature is computed as:
\begin{align}
    \frac{\partial L}{\partial F_p} = \sum_{\mathcal{P} \in \mathcal{U}} w_k \cdot \frac{\partial L}{\partial F_{(u, v)}}
\end{align}
where $k \in \{ 1, 2, 3, 4 \}$ is the index of $p_k$ in the $\mathcal{P}$. We compute the index lookup table during the forward process and save it for usage in backward. On the other hand, the gradient to $u \in \mathcal{U}$ is computed via:
\begin{align}
    \frac{\partial L}{\partial u_p} = \frac{\partial L}{\partial F_{(u, v)}} \cdot \sum_{\mathcal{P} \in \mathcal{U}} \frac{\partial F_{(u, v)}}{\partial w_k} \cdot \frac{\partial w_k}{\partial u_{p_k}} 
\end{align}
where
\begin{align}
    \frac{\partial w_k}{\partial u_{p_k}} = \frac{\partial w_k}{\partial u_{p_k}} \cdot \frac{\sum_{j \neq k}^4 w_j}{\sum_{j}^4 w_j}  - \sum_{i \neq k}^4 \frac{\partial w_i}{\partial u_{p_i}} \frac{w_i}{\sum_{j}^4 w_j}
\end{align}
and
\begin{align}
    \frac{\partial F_{(u, v)}}{\partial w_k} = \sum_{c \in C} F_{(u, v)}^c,
\end{align}
$C$ is dimension of the features. Note, in practice, we don't look up each pixel pairs on $\mathcal{U}$ to compute the sum, instead, we store the lookup and weight $w_k$ for each pixel (maximum 10 quadrilaterals per pixel) in the surface coordinate to reduce computation costs.

We implement the function using CUDA and packed as a PyTorch auto-gradient function, which are potentially useful for future projects, \eg texture extraction.

\begin{figure*}[t]
\centering
\includegraphics[width=\textwidth]{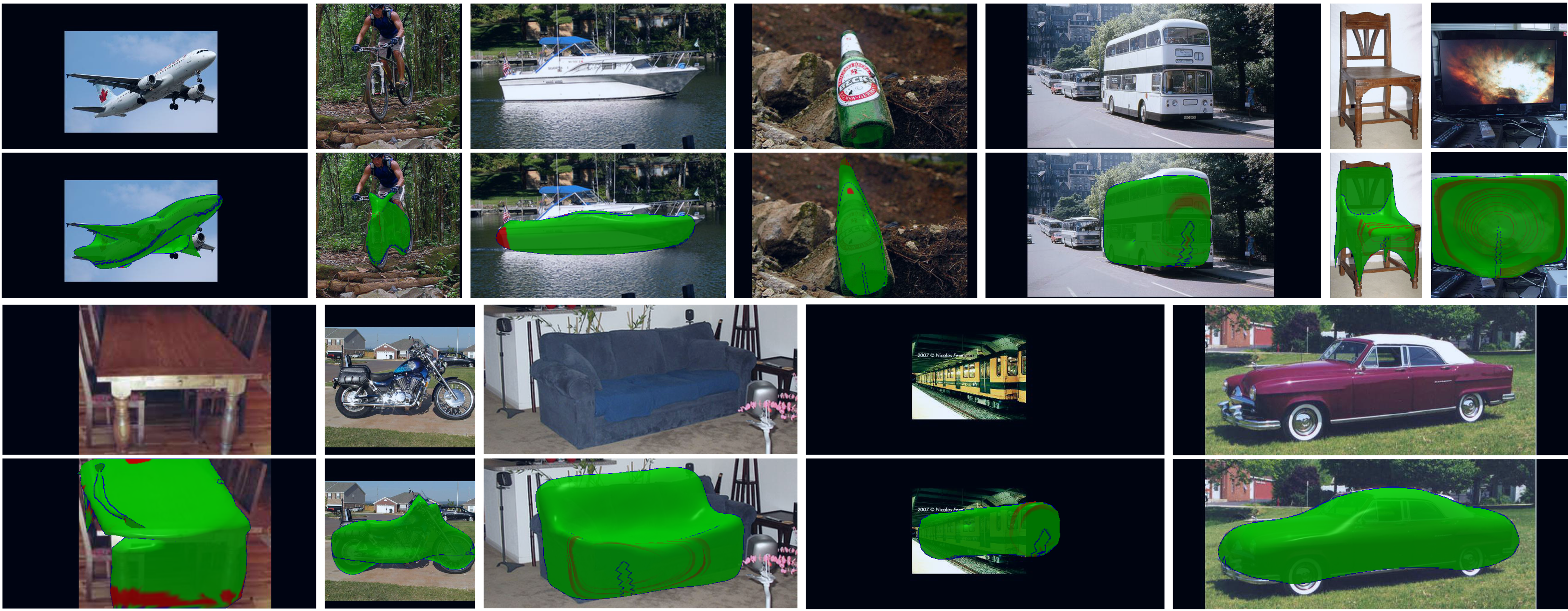}
\caption{Qualitative results of all 12 categories on PASCAL3D+ dataset \cite{pascal3dp}.}
\label{fig:qualitative-pascal3d}
\end{figure*}

\begin{figure*}[t]
\centering
\includegraphics[width=\textwidth]{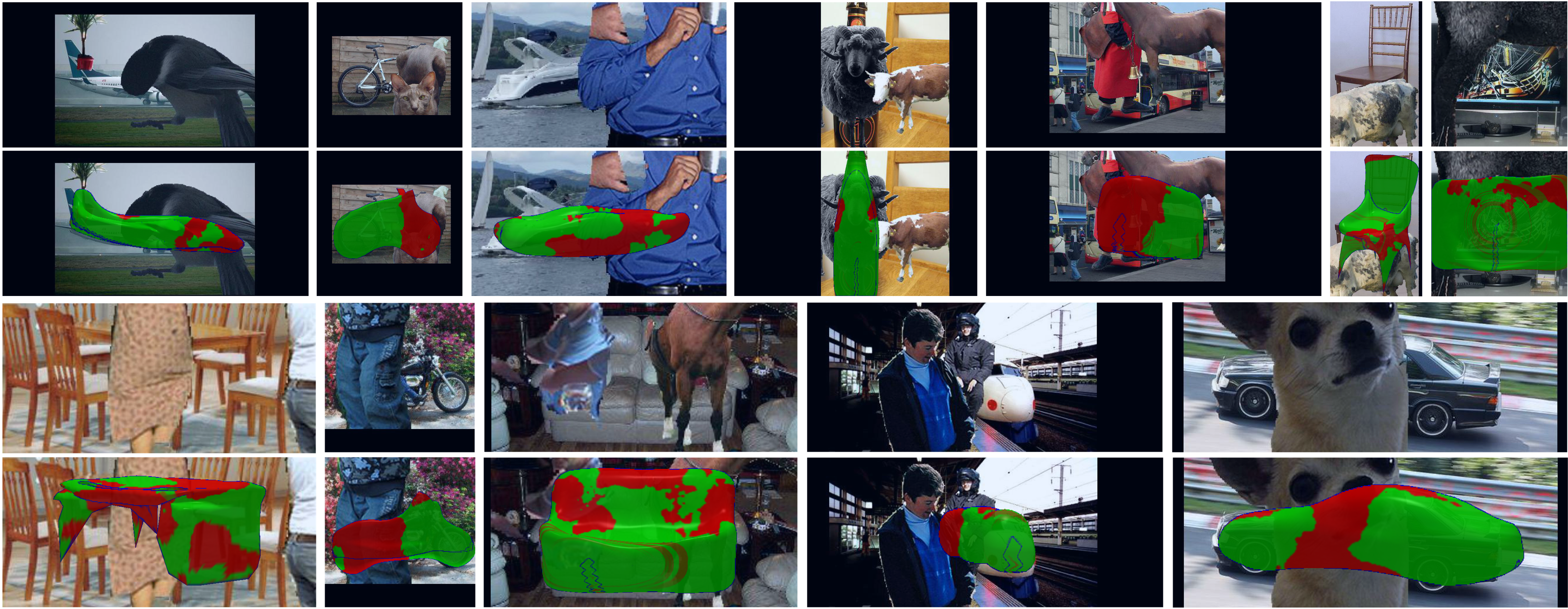}
\caption{Qualitative results of all 12 categories on Occluded PASCAL3D+ dataset \cite{wang2020robust}.}
\label{fig:qualitative-pascal3d-occ}
\end{figure*}

\subsection{Multi-Task Mask RCNN}

To compare DMNT with traditional multi-head network architectures, we extend a Mask R-CNN model \cite{He_2017_ICCV} with a pose estimation head. More specifically, the MT Mask R-CNN model consists of three heads, a classification head, a pose estimation head, and a mask segmentation head. We adopt the classification head and mask segmentation head from the official PyTorch \cite{paszke2019pytorch} implementation. We follow the loss functions defined in \cite{He_2017_ICCV}. To provide the ground-truth masks for Mask R-CNN, we compute the projection of the known CAD models given the annotated principal points and 3D poses. In terms of the pose estimation head, we follow \cite{tulsiani2015viewpoints,mousavian20173d} that formulate the pose estimation as a classification task. We follow the implementations in \cite{zhou2018starmap} to reproduce the results. Formally, the MT Mask R-CNN is end-to-end supervised by a multi-task loss given by
\begin{align}
    \mathcal{L} = \mathcal{L}_\text{cls} + \mathcal{L}_\text{box} + \mathcal{L}_\text{mask} + \mathcal{L}_\text{pose}
\end{align}

\section{Additional Qualitative Results}

In this section, we provide additional qualitative results to demonstrate the capabilities of DMNT under various settings.

\paragraph{PASCAL3D+.} We visualize the predictions of DMNT on PASCAL3D+ dataset \cite{pascal3dp} in Figure \ref{fig:qualitative-pascal3d}. By learning a 3D deformable neural representation, DMNT solve multiple tasks from a holistic perspective. As we can see from Figure \ref{fig:qualitative-pascal3d}, DMNT predicts accurate object poses, good amodal segmentations, as well as class labels.

\paragraph{Occluded PASCAL3D+.} We also visualize the results on Occluded PASCAL3D+ dataset \cite{wang2020robust} in Figure \ref{fig:qualitative-pascal3d-occ}. As we can see, DMNT is very robust to partial occlusion and can accurately capture the 3D pose and object boundaries from the visible part of the object.

\begin{figure*}[t]
\centering
\includegraphics[width=0.7\textwidth]{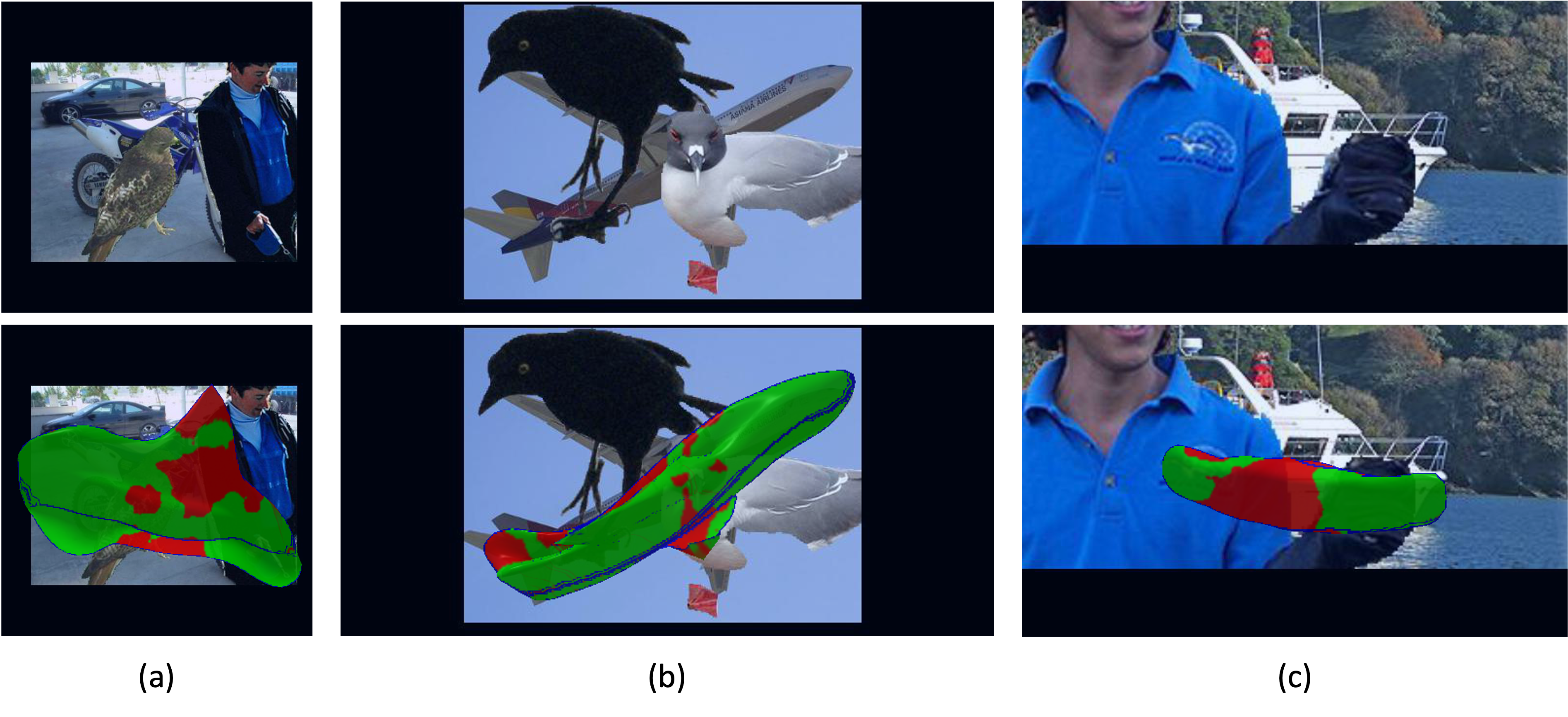}
\caption{Some failed examples from PASCAL3D+ dataset \cite{pascal3dp} and Occluded PASCAL3D+ dataset \cite{wang2020robust}.}
\label{fig:failed}
\end{figure*}

\paragraph{Failed examples.} To investigate the limitations of our model, we also looked into some failed cases from PASCAL3D+ \cite{pascal3dp} and Occluded PASCAL3D+ dataset \cite{wang2020robust}, which are shown in Figure \ref{fig:failed}. In Figure \ref{fig:failed}(a), DMNT failed to predict accurate amodal segmentation because the wheels of the car in the background resemble the wheels of the motorbike. In Figure \ref{fig:failed}(b), the novel parts of the aeroplane, i.e., wings, are largely dominated by the occluders, and DMNT failed to predict good object pose. DMNT couldn't predict good object boundaries in Figure \ref{fig:failed}(c) since this boat has a different shape from the known CAD models in PASCAL3D+ and the body of the boat is heavily occluded.

\end{document}